\documentclass{article}

\usepackage[preprint]{neurips_2024}

\usepackage[utf8]{inputenc} % allow utf-8 input
\usepackage[T1]{fontenc}    % use 8-bit T1 fonts
\usepackage{hyperref}       % hyperlinks
\usepackage{url}            % simple URL typesetting
\usepackage{booktabs}       % professional-quality tables
\usepackage{amsfonts}       % blackboard math symbols
\usepackage{nicefrac}       % compact symbols for 1/2, etc.
\usepackage{microtype}      % microtypography
\usepackage{xcolor}         % colors

\usepackage{amsfonts,amsmath,amssymb,amsthm,mathtools}
\usepackage{natbib}
\usepackage{booktabs}
\usepackage{amsmath}
\usepackage{amssymb}
\usepackage{verbatim}
\usepackage{geometry}
\usepackage{subcaption}
\usepackage{tikz}
\usepackage{pgfplots}
\usepackage{pgf}
\usepackage{xcolor}
\usepackage{adjustbox}
\usepackage{wrapfig}

\usepackage{multicol}

\usepackage{amsmath,amsfonts,bm,amsthm,amssymb}

\newcommand{\rom}[1]{\uppercase\expandafter{\romannumeral #1\relax}}

\theoremstyle{definition}

\newtheorem*{eg}{Example}

\def\eqref#1{equation~\ref{#1}}
\def\1{\bm{1}}

\DeclareMathAlphabet{\mathsfit}{\encodingdefault}{\sfdefault}{m}{sl}
\SetMathAlphabet{\mathsfit}{bold}{\encodingdefault}{\sfdefault}{bx}{n}

\def\gJ{{\mathcal{J}}}

\newcommand{\E}{\mathbb{E}}

\newcommand{\KL}{D_{\mathrm{KL}}}

\pgfplotsset{compat=1.16}
\usepgfplotslibrary{external}
\usepgfplotslibrary{fillbetween}
\tikzstyle{every plot}=[prefix=plots/]

\usetikzlibrary{automata}
\usetikzlibrary{arrows,automata}
\usetikzlibrary{arrows.meta}
\usetikzlibrary{backgrounds}
\usetikzlibrary{cd}
\usetikzlibrary{decorations.markings}
\usetikzlibrary{decorations.pathmorphing}
\usetikzlibrary{intersections}
\usetikzlibrary{positioning}
\usetikzlibrary{math}

\definecolor{econblue}{HTML}{076FA1} % #076FA1
\definecolor{econred}{HTML}{ED1F21} % #ED1F21

\definecolor{b1}{HTML}{00526D} % #00526D
\definecolor{b2}{HTML}{00A4DC} % #00A4DC
\definecolor{b3}{HTML}{70D0F6} % #70D0F6

\definecolor{b4}{HTML}{0E1D24} % #0E1D24
\definecolor{b5}{HTML}{1B3758} % #1B3758
\definecolor{b6}{HTML}{3F7CA4} % #3F7CA4
\definecolor{b7}{HTML}{53A6E5} % #53A6E5

\definecolor{l1}{HTML}{973d4c} % #973d4c
\definecolor{l2}{HTML}{AC8B96} % #AC8B96
\definecolor{l3}{HTML}{30c1d3} % #30c1d3
\definecolor{l4}{HTML}{076FA1} % #076FA1

\definecolor{c1}{HTML}{F2836B} % #F2836B
\definecolor{c2}{HTML}{F2BFAC} % #F2BFAC
\definecolor{c3}{HTML}{6394BF} % #6394BF
\definecolor{c4}{HTML}{314259} % #314259
\definecolor{c5}{HTML}{3B5E8C} % #3B5E8C

\title{Improving LLM-Generated Code Quality with GRPO}

\author{%
  Maxime Robeyns \\
  University of Bristol\thanks{iGent IA}\\
  \texttt{ez18285@bristol.ac.uk} \\
  \And
  Laurence Aitchison \\
  University of Bristol \\
  \texttt{laurence.aichison@bristol.ac.uk} \\
}

\newcommand{\rlef}{shojaee2023ppocoder,Dou2024StepCoder,Zhang2025PRLCoder,Dai2025Process,gehring2025rlef,Yang2024ACECode,Sorokin2025Iterative,Le2022CodeRL,Liu2024RLTF}
\newcommand{\codequality}{mccabe1976complexity, halstead1977elements, chidamber1994metrics, fowler1999refactoring, martin2008clean, lanza2007object}

\begin{document}

\maketitle

\begin{abstract}
Large Language Models (LLMs) are gaining widespread use for code generation. Recent training procedures use execution feedback as a reward signal, typically focusing on the functional correctness of the code, using unit test pass rate as a reward signal. However, this reward signal fails to capture notions of maintainability, quality and safety of the code produced. We address this under-explored area and develop a comprehensive library to quantify various aspects of code quality, and use it as a reward in GRPO. We find GRPO increases code quality according to this measure, which is confirmed by expert, blinded human annotators.
\end{abstract}

\section{Introduction}%
\label{sec:intro}

An increasing proportion of the world's software is being generated by LLMs.
However, if LLM code generation is to continue gaining trust and adoption,
understanding and improving the quality of LLM generated code is essential: it
is quite possible for a ``vibe-coded'' software component to
become unmaintainable by LLMs or expert software engineers, be open to critical
security vulnerabilities, or waste vast amounts of energy by e.g.\ using a
quadratic algorithm where a linear one is available.

Recently, these LLM coding systems have been trained using \textit{execution
  feedback} as a reward signal \citep[e.g.][]{\rlef}. An example of an execution
feedback setting is to take a coding specification or question (e.g.\ ``write a
function that generates Fibonacci numbers'') paired with a number of tests.
After the LLM generates code to solve the problem, we check whether that code
passes the tests.  If the code does pass the tests, then that is counted as a
correct response and rewarded e.g.\ in a GRPO \citep{shao2024deepseekmath}
pipeline.  If the code does not pass one or more of the tests, then that is
counted as an incorrect response and is penalized.

However, these rewards lack any notion of software quality, including:
\begin{itemize}
  \item Maintainability: how easy is the software to keep working with and modify in the future
  \item Security: how well protected the code is against vulnerabilities
  \item Reliability: how effectively the software ensures availability, fault tolerance, and recoverability
  \item Performance: how well the generated code uses resources
\end{itemize}
In order to improve the usefulness and adoption of LLM generated code, we must
look beyond merely rewarding functional correctness, but include incentives to
produce code with good style and quality attributes, which make the code more
readable, easier to work with and extend.

We thus sought to introduce a new family of rewards for e.g.\ GRPO that care
about code-quality. That of course required us to programmatically quantify
code-quality. Thankfully, quantifying code-quality is an area that has been
well studied in the Computer Science literature \citep[e.g.][]{\codequality}, so
there are many reasonable automated metrics which capture many of the four broad
aspects listed above.
As a concrete starting point, considered the list of automated source code
quality
measures\footnote{https://www.it-cisq.org/cisq-files/pdf/cisq-weaknesses-in-ascqm.pdf}
from the Consortium for Information \& Software Quality (CISQ)
\citep{cisqstandards}. We developed a Python library,
\verb|codequal_analyzer|, which implements analyzers for many of the common code
weaknesses identified in the CISQ standards, and in turn evaluate the quality of
Python code. This includes the ability to map identified code flaws to their
Common Weakness Enumeration (CWE) ID, and return scalar code quality scores
suitable for use within an RL pipeline.
Next, we investigated whether incorporating this reward in a GRPO pipeline would improve
code-quality in practice, relative to a control GRPO pipeline with no
code-quality reward.
We found that it did, both as measured by our code-quality metric, and by human
annotators who were presented with answers from each of the resulting models,
and asked to pick the one with better code quality. Of course, these annotators
were blinded in the sense that the were not told which model each answer came
from.
Importantly, we also found that the model trained with a code-quality reward
both performed as well or even better than the baseline model (measured using correctness, i.e.\ whether the code passes the
tests), while also producing code of a shorter length on average than the
baseline model. This means that at deployment-time, this intervention yields
improved code quality while incurring no additional generation costs.

Our contributions are:
\begin{enumerate}
    \item A comprehensive library, \verb|codequal_analyzer| which captures notions
        of code-quality as defined by CISQ, mapping issues back to CWE IDs, and giving
        scores suitable for use within an RL pipeline\footnote{\url{https://github.com/MaximeRobeyns/codequal_analyzer}}.
    \item Demonstrating that GRPO with a code-quality reward can indeed improve
        the quality of generated code, as evaluated by expert human annotators.
\end{enumerate}

\section{Related Work}%
\label{sec:related_work}

There are a large number of papers using execution feedback to train LLMs to write code that passes tests \citep[e.g.][]{\rlef}.
However, to our knowledge there is as of yet no work that combines this approach with reward terms to encourage improved code quality (our key contribution in this paper).

At the same time, there is a classical literature in computer science on code quality, including how to understand, improve and quantify it \citep[e.g.][]{\codequality}.
However, to our knowledge there is as of yet no work that takes these metrics for code quality and uses them as a reward signal for training LLMs to produce higher-quality code.

Our \verb|codequal_analyzer| library for assessing code-quality is designed to follow \citet{cisqstandards}, and uses a number of pre-existing libraries \citep{pylint,radon,vulture,bandit,safety,mypy}, along with a considerable number of ``analyzers'' written from scratch (see Sec.~\ref{sec:measuring_code_quality} for details).
Importantly, these libraries have not, to our knowledge, been integrated into a comprehensive framework which produces a single number suitable for use as a reward in RL.

\section{Methods}%
\label{sec:method}

We use GRPO to improve the coding ability of various open-source LLMs.
Such a pipeline involves making multiple choices, including the dataset and reward design, which we describe below.

\begin{figure}[tbp]
  \begin{subfigure}[b]{0.5\textwidth}
    \includegraphics[width=\textwidth]{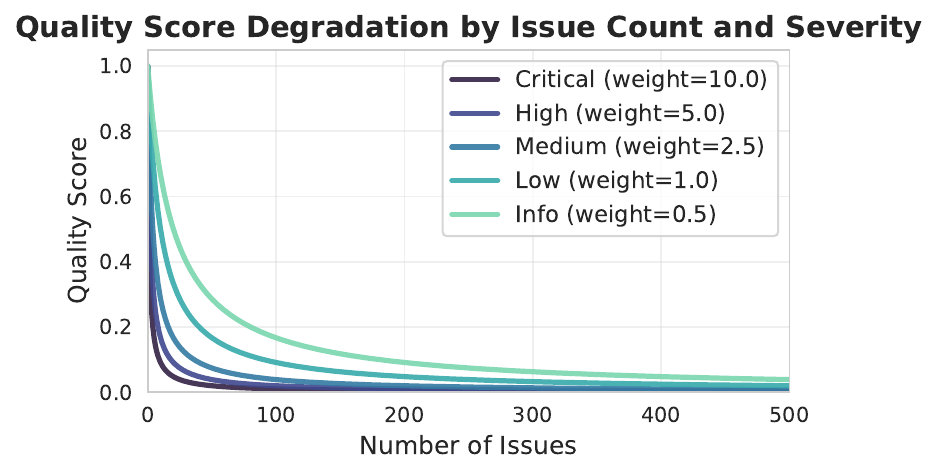}
  \end{subfigure}
  \hfill
  \begin{subfigure}[b]{0.5\textwidth}
    \includegraphics[width=\textwidth]{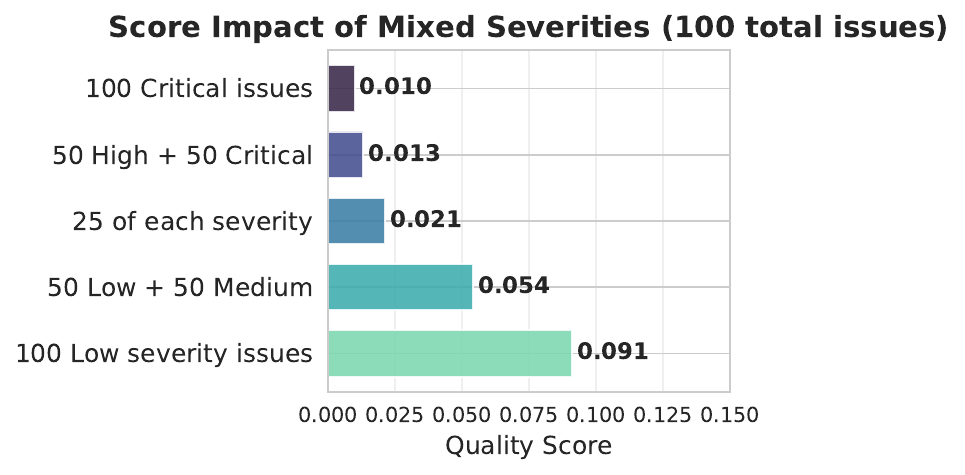}
  \end{subfigure}
  \caption{Code quality score evolution over number of issues and issue severity
    level}
  \label{fig:reward_evolution}
\end{figure}

\subsection{Measuring Code Quality in RL Pipelines}
\label{sec:measuring_code_quality}

We began by implementing a comprehensive library for evaluating the quality of
Python code.  We started with the CISQ Standards \citep{cisqstandards}.
While comprehensive, these standards are in natural-language form, so not come
with an official implementation.  As such, we combined a number of existing
libraries, such as Pylint \citep{pylint}, Radon \citep{radon}, MyPy \citep{mypy}
and others that are able to detect code quality issues, while writing
from-scratch a number of ``analyzers'' to detect issues that are missed by these
general tools. We taxonomize the aspects of code quality that our analyzer
picks up on in Table~\ref{tab:code-quality-issues}.

\begin{table}[htbp]
\centering
\caption{Taxonomy of Code Quality Issues Detected by CISQ Analyzer}
\label{tab:code-quality-issues}
\begin{adjustbox}{max width=\linewidth, center}
\begin{tabular}{@{}p{3.5cm}p{7cm}l@{}}
\toprule
\textbf{Category} & \textbf{Example Issues} & \textbf{Analyzer} \\
\midrule
\multicolumn{3}{@{}l}{\textbf{Maintainability}} \\
\addlinespace
\textit{Code Complexity} & Excessive cyclomatic complexity & Radon \citep{radon} \\
 & Functions with high complexity scores & Xenon \citep{xenon} \\
 & Classes with overly complex methods & \\
\addlinespace
\textit{Dead Code} & Unused functions, methods, and variables & Vulture \citep{vulture} \\
 & Unused class definitions and imports & \\
\addlinespace
\textit{Code Structure} & Excessive function arguments & Pylint \citep{pylint} \\
 & Too many instance attributes & \\
 & Large files (>1000 LOC) & \\
 & Excessive branches/returns & \\
\addlinespace
\textit{Style \& Documentation} & Missing docstrings & Pylint \\
 & Poor naming conventions & \\
\midrule
\multicolumn{3}{@{}l}{\textbf{Security}} \\
\textit{Code Injection} & Shell injection (os.system) & Bandit \citep{bandit} \\
 & Unsafe subprocess calls & \\
 & Command injection risks & \\
\addlinespace
\textit{Unsafe Data Handling} & Insecure deserialization (pickle, YAML) & Bandit \\
 & Insecure XML parsing & \\
\addlinespace
\textit{Cryptography} & Weak hash algorithms (MD5, SHA1) & Bandit \\
 & Insecure random generation & \\
 & Hard-coded secrets & \\
\addlinespace
\textit{Dependencies} & Known vulnerable packages & \emph{custom} \\
\midrule
\multicolumn{3}{@{}l}{\textbf{Performance}} \\
\addlinespace
\textit{String Operations} & String concatenation in loops (+, +=) & \emph{custom} \\
\addlinespace
\textit{Resource Utilization} & Resource-intensive loop operations & \emph{custom} \\
 & Growing data structures in loops &  \\
 & Network/file I/O in loops & \\
\addlinespace
\textit{Data Structures} & Excessive class attributes & \emph{custom} \\
 & Deeply nested structures & \\
 & Large dictionaries & \\
\midrule
\multicolumn{3}{@{}l}{\textbf{Reliability}} \\
\addlinespace
\textit{Exception Handling} & Bare/empty except clauses & \emph{custom} \\
 & Overly broad exception catching & \\
 & Missing resource cleanup & \\
\addlinespace
\textit{Concurrency} & Lock ordering issues & \emph{custom} \\
 & Missing lock releases & \\
\addlinespace
\textit{Infinite Loops} & Missing exit conditions & \emph{custom} \\
 & Unchanging loop counters & \\
 & While True without breaks & \\
\addlinespace
\textit{Type Safety} & Type inconsistencies & Mypy \citep{mypy} \\
 & Missing annotations & \\
 & Incorrect argument/return types & \\
\bottomrule
\end{tabular}
\end{adjustbox}
\end{table}

We collect these different code quality heuristics into a single
\verb|codequal_analyzer| library. Given a path to a directory of code to analyze,
the main analysis function runs all the analysers for all characteristic groups
(maintainability, security, performance, reliability) in parallel on the code,
accumulating any found issues. These findings usually include a mapping to the
CISQ CWE ID for categorization, and also include an assessment of the issue's
severity into the set $\mathcal S=\{\text{info},\text{low},\text{medium},\text{high},\text{critical}\}$.

To obtain a numerical score which to train a model, we aggregate the findings as
follows. First, we define the following weightings for each severity level,
reflecting the relative importance to place on each type of issue identified:
\[
\begin{alignedat}{4}
w_{\text{info}}     &= 0.5, &\quad
w_{\text{low}}      &= 1.0, &\quad
w_{\text{medium}}   &= 2.5, &\quad
w_{\text{high}}     &= 5.0, &\quad
w_{\text{critical}} &= 10.0.
\end{alignedat}
\]

Then, we let $N_{s}$ be the number of findings at severity level $s$ and
calculate the weighted sum across severity levels
\[
W = \sum_{s\in\mathcal S} w_s\,N_{s},
\]
following which we obtain a score between $0$ and $1$ using the following
formula that decays with the weighted finding count, which is visually
illustrated in Figure~\ref{fig:reward_evolution}:
\[
  r_{\text{quality}} = \frac{1}{1+W}.
\]

% https://www.it-cisq.org/standards/code-quality-standards/
% https://www.iso.org/obp/ui/en/#iso:std:iso-iec:5055:ed-1:v1:en
% https://www.it-cisq.org/cisq-files/pdf/cisq-weaknesses-in-ascqm.pdf

\subsection{Policy Optimization Algorithm}

For completeness, we describe the Group Relative Policy Optimization
\citep{shao2024deepseekmath} algorithm we use to train the model, and
modifications from subsequent papers.
The core idea behind GRPO is to sample
multiple candidate outputs for a given query, and use their relative rewards to
estimate advantages for policy updates. For each query sampled from the dataset
$q \sim P(Q)$, GRPO samples a group $G$ of outputs $o_{1}, \ldots, o_{G}$ using
a behaviour policy $\pi_{\theta_{\text{old}}}$, corresponding to a previous
iteration of the policy model, and updated periodically. The policy $\pi_{\theta}$
is updated by maximizing the following objective:
\begin{align}
  \label{eq:grpo_objective}
  \gJ_{\text{GRPO}}&(\theta) = \E_{q\sim P(Q), \{o_{i}\}^{G}_{i=1}\sim \pi_{\theta_{\text{old}}}(O\vert q)} \\
  &\frac{1}{G}\sum^{G}_{i=1}\frac{1}{\vert o_{i}\vert}\sum^{\vert o_{i}\vert}_{t=1}\min\bigg(\left[c_{i,t}(\theta)\hat{A}_{i,t},\,\text{clip}\big(c_{i,t}(\theta),1-\epsilon_{\text{low}},1+\epsilon_{\text{high}}\big)\hat{A}_{i,t}\right] - \beta\KL\left[\pi_{\theta} \Vert \pi_{\text{ref}}\right]\bigg), \nonumber
\end{align}
where
\[
  c_{i,t}(\theta) = \frac{\pi_{\theta}(o_{i,t} \vert q, o_{i,<t})}{\pi_{\theta_{\text{old}}}(o_{i,t}\vert q, o_{i,<t})},\quad \KL\left[\pi_{\theta} \vert \pi_{\text{ref}}\right] = \frac{\pi_{\text{ref}}(o_{i,t} \vert q, o_{i, <t})}{\pi_{\theta}(o_{i,t} \vert q, o_{i,<t})} - \log \frac{\pi_{\text{ref}}(o_{i,t} \vert q, o_{i, <t})}{\pi_{\theta}(o_{i,t} \vert q, o_{i<t})} - 1,
\]
with clipping hyperparameters set to $\epsilon_{\text{low}} = 0.2$, $\epsilon_{\text{high}} = 0.28$ following \citep{yuDAPOOpenSourceLLM2025},
 KL penalty strength coefficient $\beta = 0.001$ controlling stability and exploration
 \citep{schulmanProximalPolicyOptimization2017}, and $\hat{A}_{i,t}$ being the
 group-relative advantage replicated for each token in the trajectory $o_{i}$
 \[
   \hat{A}_{i,t} = \frac{r_{i} - \text{mean}(\{r_{1}, r_{2}, \ldots, r_{G}\})}{\text{std}(\{r_{1}, r_{2}, \ldots, r_{G}\})}.
 \]

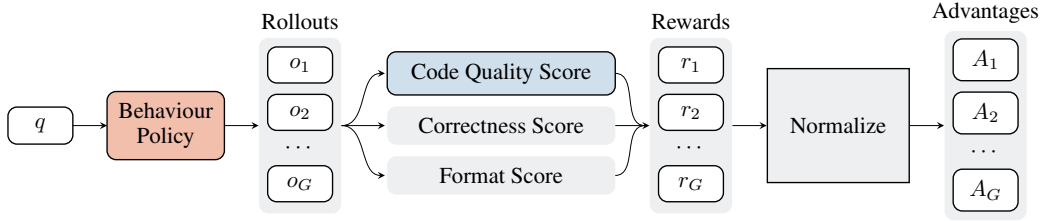
\begin{figure}[t]
  \centering
  \pgfdeclarelayer{background}
\pgfdeclarelayer{foreground}
\pgfsetlayers{background,main,foreground}
\adjustbox{width=1.0\linewidth}{
   \begin{tikzpicture}[scale=1,
     block/.style={draw, align=center, minimum width=1.0cm, inner sep=5pt, outer
     sep=0pt, rounded corners, semithick}
     ]

     \node (a) at (0, 0) [block] {$q$};%

     \node (b) [right=15pt of a, block, fill=c2] {Behaviour \\ Policy};%

     \draw[-stealth] (a) -- (b);%

     % Rollouts ------------------------
     \node (c) [right=15pt of b, fill=b4!7, minimum height=77pt, minimum
     width=36pt, rounded corners, label={Rollouts}] {};%
     \node (d) [above=-20pt of c, block, fill=white] {$o_{1}$};%
     \node (e) [below=5pt of d, block, fill=white] {$o_{2}$};%
     \node (f) [below=2pt of e] {$\cdots$};%
     \node (g) [below=2pt of f, block, fill=white] {$o_{G}$};%

     \draw[-stealth] (b) -- (c);%

     % Reward components ---------------
     \node (h) [right=20pt of c, block, fill=b4!7, draw=none, minimum width=100pt] {Correctness Score};%
     \node (i) [above=5pt of h, block, fill=c3!30, minimum width=100pt] {Code Quality Score};%
     \node (j) [below=5pt of h, block, fill=b4!7, draw=none, minimum width=100pt] {Format Score};%

     % Reward arrows
     \draw[-stealth] (c.east) -- ([xshift=-1pt]h.west);%
     \draw[-stealth] (c.east) to[out=0, in=180, rounded corners=3pt] ([xshift=-1pt]i.west);%
     \draw[-stealth] (c.east) to[out=0, in=180, rounded corners=3pt] ([xshift=-1pt]j.west);%

     % Rewards
     \node (k) [right=15pt of h, fill=b4!7, minimum height=77pt, minimum
     width=36pt, rounded corners, label={Rewards}] {};%
     \node (l) [above=-20pt of k, block, fill=white] {$r_{1}$};%
     \node (m) [below=5pt of l, block, fill=white] {$r_{2}$};%
     \node (n) [below=2pt of m] {$\cdots$};%
     \node (o) [below=2pt of n, block, fill=white] {$r_{G}$};%

     \draw[-stealth] (h.east) -- (k.west);%
     \draw[-] (i.east) to[out=0, in=180, rounded corners=3pt] ([xshift=-2pt]k.west);%
     \draw[-] (j.east) to[out=0, in=180, rounded corners=3pt] ([xshift=-2pt]k.west);%

     % Normalization
     \node (p) [right=15pt of k, fill=b4!7, draw, semithick, minimum
     height=50pt, inner sep=10pt] {Normalize};%
     \draw[-stealth] (k) -- (p);%

     % Advantages
     \node (q) [right=15pt of p, fill=b4!7, minimum height=83pt, minimum
     width=36pt, rounded corners, label={Advantages}] {};%
     \node (r) [above=-22pt of q, block, fill=white] {$A_{1}$};%
     \node (s) [below=5pt of r, block, fill=white] {$A_{2}$};%
     \node (t) [below=2pt of s] {$\cdots$};%
     \node (u) [below=2pt of t, block, fill=white] {$A_{G}$};%

     \draw[-stealth] (p) -- (q);%
   \end{tikzpicture}
}
  \caption{\label{fig:grpo_rewards} GRPO advantage calculation. In our experiments, we ablate the
    code quality score to quantify the benefit of including it.}
\end{figure}

We illustrate this in Figure~\ref{fig:grpo_rewards}.

\subsection{Reward Design for Coding Tasks}

Our rewards for each rollout $r_{i}$ combines three components: a very simple
format reward $r_{i, \text{format}}$ (which ensures the code can be parsed
correctly), a code correctness reward $r_{i, \text{correct}}$ (which ensures the
code functions correctly) as well as our code quality reward signal
$r_{i, \text{quality}}$.
Each of these range from 0 to 1.

The rewards for each rollout are linearly combined, with a slight emphasis on
the code quality over the other components:
\begin{equation}
  \label{eq:reward_with_qual}
  r_{i} = \frac{2}{10}r_{i, \text{format}} + \frac{3}{10}r_{i, \text{correct}} + \frac{5}{10}r_{i, \text{quality}}.
\end{equation}

\textbf{Format Reward}
Recent work has raised concerns that much of the
performance improvement in RL on LLMs for coding can be explained by the LLM
learning to generate code in the context of the specific prompts and tool format
used, without much change to the code generation properties of the model itself
\citep{shao2025spurious,chandakIncorrectBaselineEvaluations2025}. To avoid this
conflation and focus on the code quality, we simply prompt the model to output
the solution in a Python markdown code block. We reward the model for
one well formatted code block, and penalise multiple or incomplete code blocks.

\textbf{Correctness Reward.}
The continuous correctness reward measures the held-out unit test pass rate,
measuring whether the final code is functional and correct. A score of 0.0
indicates no tests passed, and 1.0 indicates all tests passed.

\textbf{Code Quality Reward.}
This is the average score produced by \verb|codequal_analyzer|. This ranges from 0
to 1, and does not rely on any `labels' (i.e. unit test suite).

\begin{figure}[tbp]
  \begin{subfigure}[b]{0.45\textwidth}
    \includegraphics[width=1.05\textwidth]{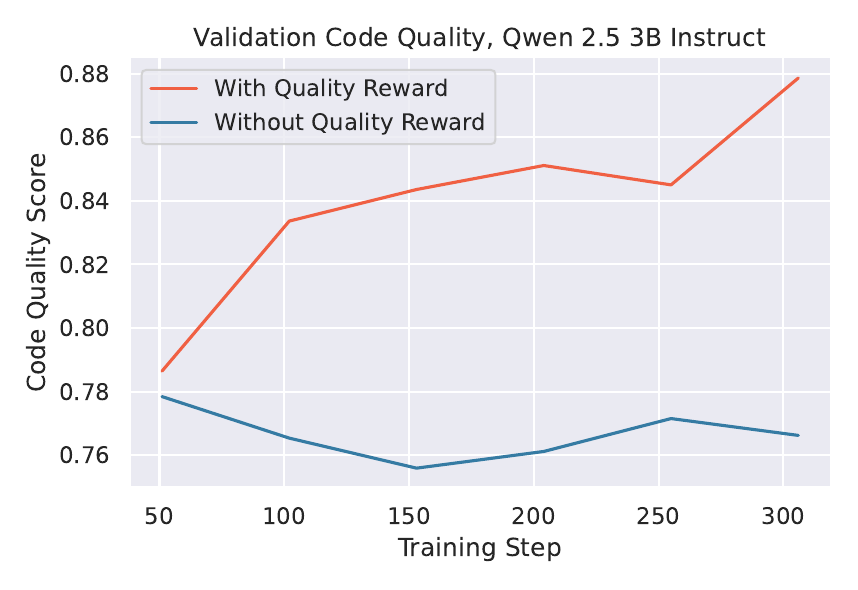}
  \end{subfigure}
  \begin{subfigure}[b]{0.45\textwidth}
    \includegraphics[width=1.05\textwidth]{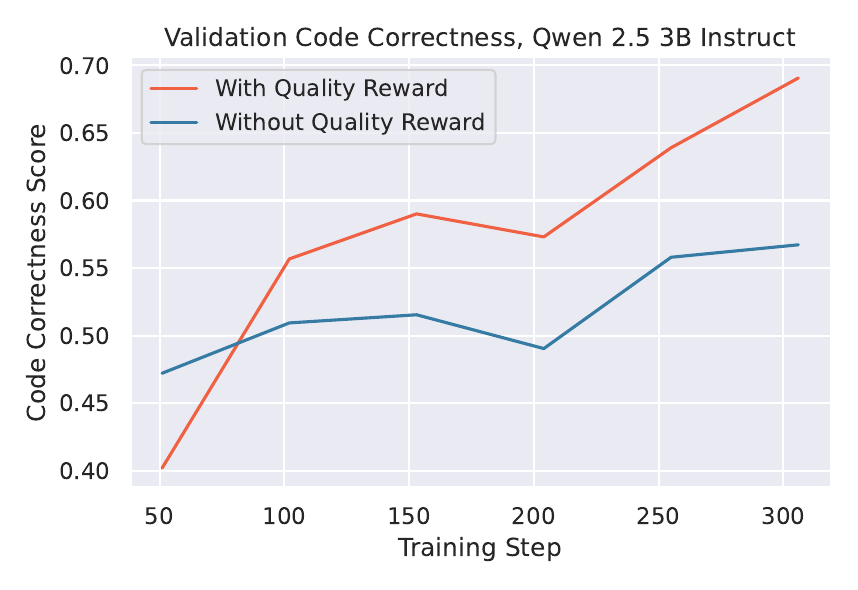}
  \end{subfigure}
  \caption{Validation code quality score and correctness throughout Qwen 2.5 3B
    training. The model trained with the code quality reward sees an appreciable
  10\% improvement in quality, while maintaining similar or better code correctness.}
  \label{fig:quality_correctness}
\end{figure}

\subsection{Synthetic Dataset Generation}

Standard benchmarks like MBPP \citep{austin2021program}, HumanEval
\citep{chen2021evaluating}, and APPS \citep{hendrycksapps2021} proved to be of
limited utility for our analysis, as we found that the problems lacked the
complexity required to meaningfully differentiate between solutions using
established code quality metrics. The types of problems in these datasets are
relatively self-contained and algorithmic in nature, and do not provide
sufficient problem variation to demonstrate a number of code quality issues like
credential handling, error handling, safe use of subprocesses and so on.
Thus, we designed a synthetic data generation pipeline, which allowed us to
generate multiple unique code-editing problems which were likely to highlight
code quality issues in the models, while also being able to vary the theme,
complexity and token length of the problems through prompting and filtering.

The synthetic data generation process is as follows: we first choose a problem
category and a subcategory from lists spanning e.g.\ algorithm optimisation,
data structures, programming paradigms, error handling and so forth.  We then
give Gemini 2.5 Pro (03-25) the category and subcategory, and prompt it to
generate a problem statement. Before proceeding, we assess the conceptual
novelty of the problem given the list of previously generated problems, and
reject ones which are merely variations on previous problems, ensuring
diversity. We then generate some starter code (e.g.\ a suboptimal solution),
an ideal solution, and set of test cases. We finally iterate on the tests to
ensure they are correct and pass with the reference solution.

See Appendix~\ref{sec:problem_examples} for more detail about the dataset problems.

\section{Results}%
\label{sec:results}

\begin{table}[t]
\centering
\caption{Reward components measured on the final iteration on the held-out
  validation problems.}
\label{tab:results}
\begin{tabular}{llll}
\toprule
Model & Validation Quality & Validation Correctness & Total Reward \\
\midrule
Qwen 2.5 3B-Instruct & $0.766$ & $0.567$ & $0.603$ \\
\emph{+ quality reward training} & $\mathbf{0.878 (+0.112)}$ & $\mathbf{0.690 (+0.123)}$ & $\mathbf{0.785 (+0.182)}$ \\
\addlinespace
Llama 3.2 3B & $0.859$ & $0.601$ & $0.627$ \\
\emph{+ quality reward training} & $\mathbf{0.894 (+0.035)}$ & $\mathbf{0.609 (+0.008)}$ & $\mathbf{0.760 (+0.133)}$ \\
\addlinespace
OLMo 2 0425 1B Instruct & $0.791$ & $0.305$ & $0.410$ \\
\emph{+ quality reward training} & $\mathbf{0.864 (+0.073)}$ & $0.217 (-0.088)$ & $\mathbf{0.631 (+0.221)}$ \\
\bottomrule
\end{tabular}
\end{table}

\begin{wrapfigure}{R}{0.5\textwidth}
\includegraphics[width=0.5\textwidth]{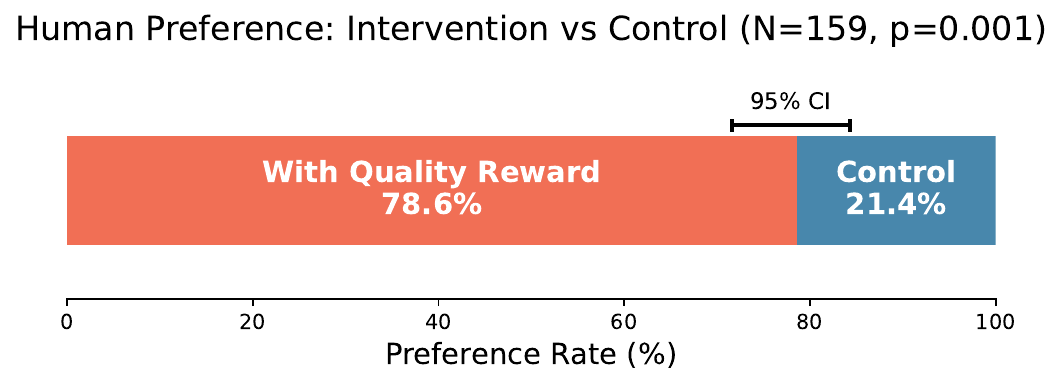}
\label{fig:human_preferences}
\caption{Human preferences of Qwen 2.5 3B output with and without the quality reward signal.}
\end{wrapfigure}

We applied GRPO on Llama 3.2 3B Instruct \citep{grattafioriLlama3Herd2024},
Qwen2.5 3B Instruct \citep{qwenQwen25TechnicalReport2025} and Olmo 2 1B
Instruct \citep{olmo2OLMo22025} with a dataset of 200 Python coding
problems generated from our synthetic data generation pipeline, and our set of
reward signals. We report the main findings in Table~\ref{tab:results}.

First, we found that models trained with and without the quality reward
component performed similarly in terms of code correctness on the held-out
set of validation coding problems, with slight improvements even observed in the
Qwen and Llama models. When considering the quality reward component, as we
might expect, the models trained with the \texttt{codequal\_analyzer}-based quality
reward component had higher code quality scores when evaluated on the held-out
validation set.

To check for reward hacking, we presented human evaluators with pairs of
solutions to validation problems generated from the models trained with and
without the quality reward. These solutions were anonymized, presented in
a random order, and the annotators were simply told to  ``\emph{Choose which
code snippet you think is of higher quality by clicking on it.}''

The human evaluators preferred the output from the model trained with the code
quality component in 78.6\% of 159 comparisons (95\% CI: 71.6\%-84.3\%, p <
0.001, binomial test). The effect size was large (Cohen's h = 0.609). The
position randomization was also effective (p = 0.812), with no significant
position bias observed (p = 0.096).

Finally here is a short example of the difference in code produced in
Figure~\ref{fig:example_code}. See Appendix~\ref{app:examples} for some more
examples. Despite this being a short problem, we can see the version from the
model trained with the quality reward signal does not include the unused
\texttt{math} import, includes type hints on the function, signature, and checks
for an early return condition which simplifies the sum and
index error handling later.

\begin{figure}[htbp]
  \centering
  \includegraphics[width=1.0\linewidth]{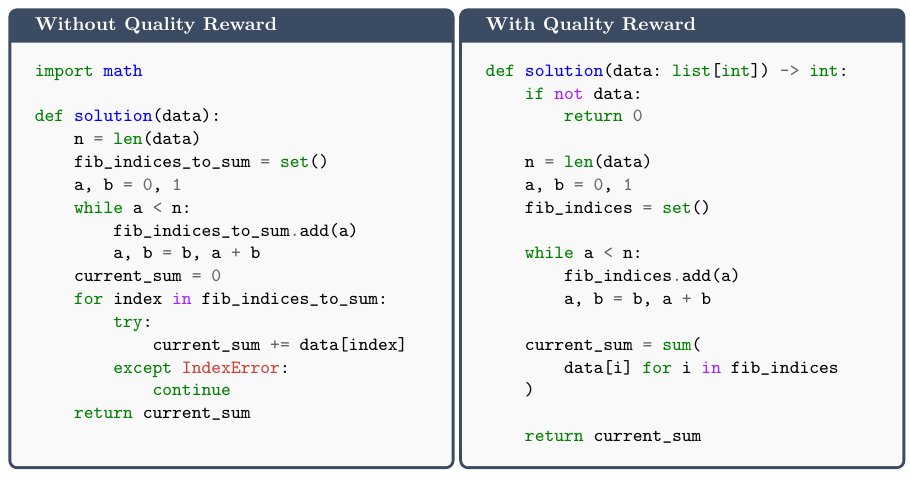}
\caption{Example code from models trained with and without the quality score.}
\label{fig:example_code}
\end{figure}

The \texttt{codequal\_analyzer} library is relatively CPU inexpensive, and executes
all the analyzers in parallel to return the quality reward score in well under
1s per rollout.
% While the execution may be offloaded on distributed compute such as
% Modal\footnote{https://modal.com/} to scale to larger batch and group sizes, we
% found this was largely unnecessary owing to the low CPU cost of running the
% analyzers and since the primary computational bottleneck during training remains.

\section{Conclusion}%
\label{sec:conclusion}

In this work, we addressed the prevalent challenge of suboptimal code quality in
Large Language Models (LLMs), which often stems from training methodologies that
prioritize execution feedback over quality considerations.
We introduced \verb|codequal_analyzer|, a novel, comprehensive library grounded in
CISQ standards, designed to quantify multiple facets of code quality—including
maintainability, security, reliability, and performance—and translate them into
a reward signal suitable for Reinforcement Learning (RL) pipelines.  By
incorporating this quality metric into a GRPO framework alongside rewards for
correctness, and utilizing a purpose-built synthetic dataset reflecting
real-world code-editing scenarios, we successfully trained LLMs to generate
higher-quality code.
Our findings indicate a significant improvement in code quality, as measured by
our automated metrics and, importantly, validated by blinded expert human
annotators without any degradation in the functional correctness of the
generated code compared to baseline models.

\bibliographystyle{plainnat}
\bibliography{refs}

\newpage
\appendix

\section{Synthetic Code Problem Examples}%
\label{sec:problem_examples}

Our dataset generation is structured across the following problem categories:
{'algorithm selection', 'array manipulation', 'custom structures', 'data structure choice',
 'decomposition', 'edge cases', 'exception handling', 'extract function', 'function composition',
 'graph algorithms', 'hash table usage', 'immutability', 'logical errors', 'loop efficiency',
 'map filter reduce', 'memoization', 'null handling', 'off by one', 'pure functions',
 'recursion patterns', 'redundant work', 'remove duplication', 'simplify conditionals',
 'tree operations', 'variable renaming'}

Each problem in the dataset contains a problem id, estimated difficulty level,
a natural language problem statement, an initial (sub-optimal) code solution, an
ideal solution and unit test cases.

Here is an example from the 'redundant work' category in the training dataset.
The problem statement is:

{
    \small \texttt{You are given a list of tasks, where each task has an ID, a category, and an initial priority. You are also given a list of operations. Each operation is of the form `('UPDATE\_PRIORITY', category\_name, new\_priority)`, indicating that all tasks belonging to `category\_name` should have their priority changed to `new\_priority`. If multiple operations target the same category, the latest operation in the list for that category determines its final priority.  Your objective is to calculate the total sum of the final priorities of all tasks after considering all operations. To optimize, first determine the definitive priority for each category affected by operations. Then, sum the priorities of all tasks, using the determined category priority if available, or the task's original priority otherwise. This avoids redundantly processing updates for tasks.}
}

The initial code that the LLM must improve is printed overleaf:
\newpage

\begin{figure}
\centering
\includegraphics[width=1.0\linewidth]{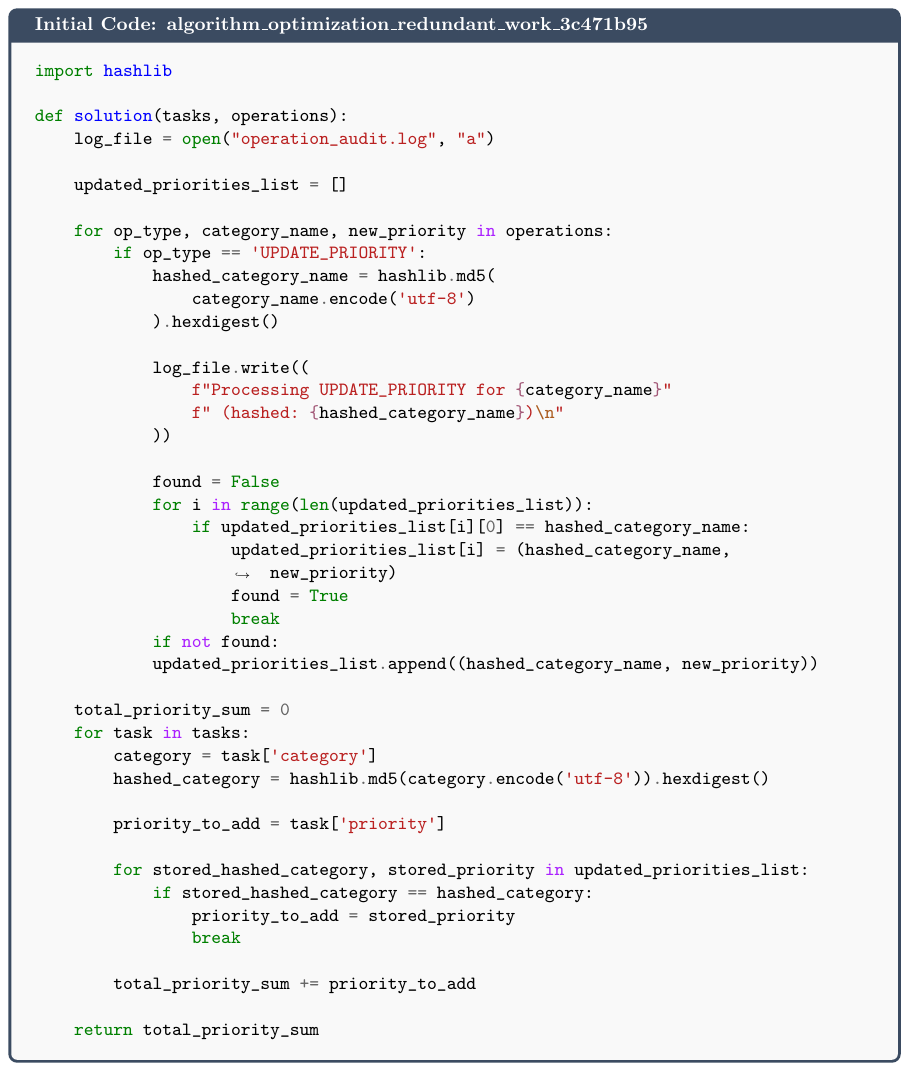}
\end{figure}

\section{More Code Examples}%
\label{app:examples}

Here are some more side-by-side code examples from Qwen 2.5 3B trained with and
without the quality reward component. Note that these may not be functionally
identical, with some potentially containing bugs.

\begin{figure}[htbp]
\centering
\includegraphics[width=1.0\linewidth]{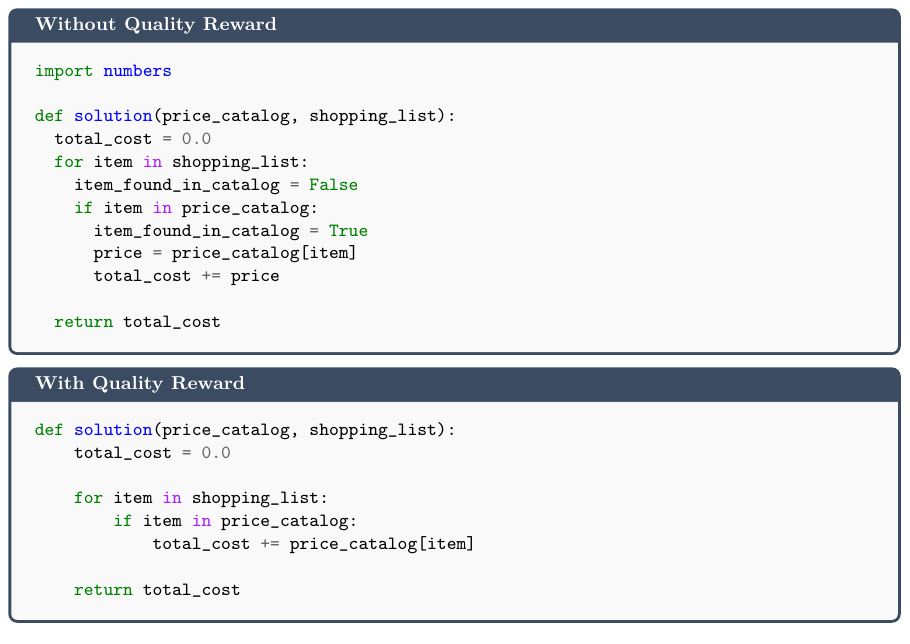}
\end{figure}

\begin{figure}[htbp]
\centering
\includegraphics[width=1.0\linewidth]{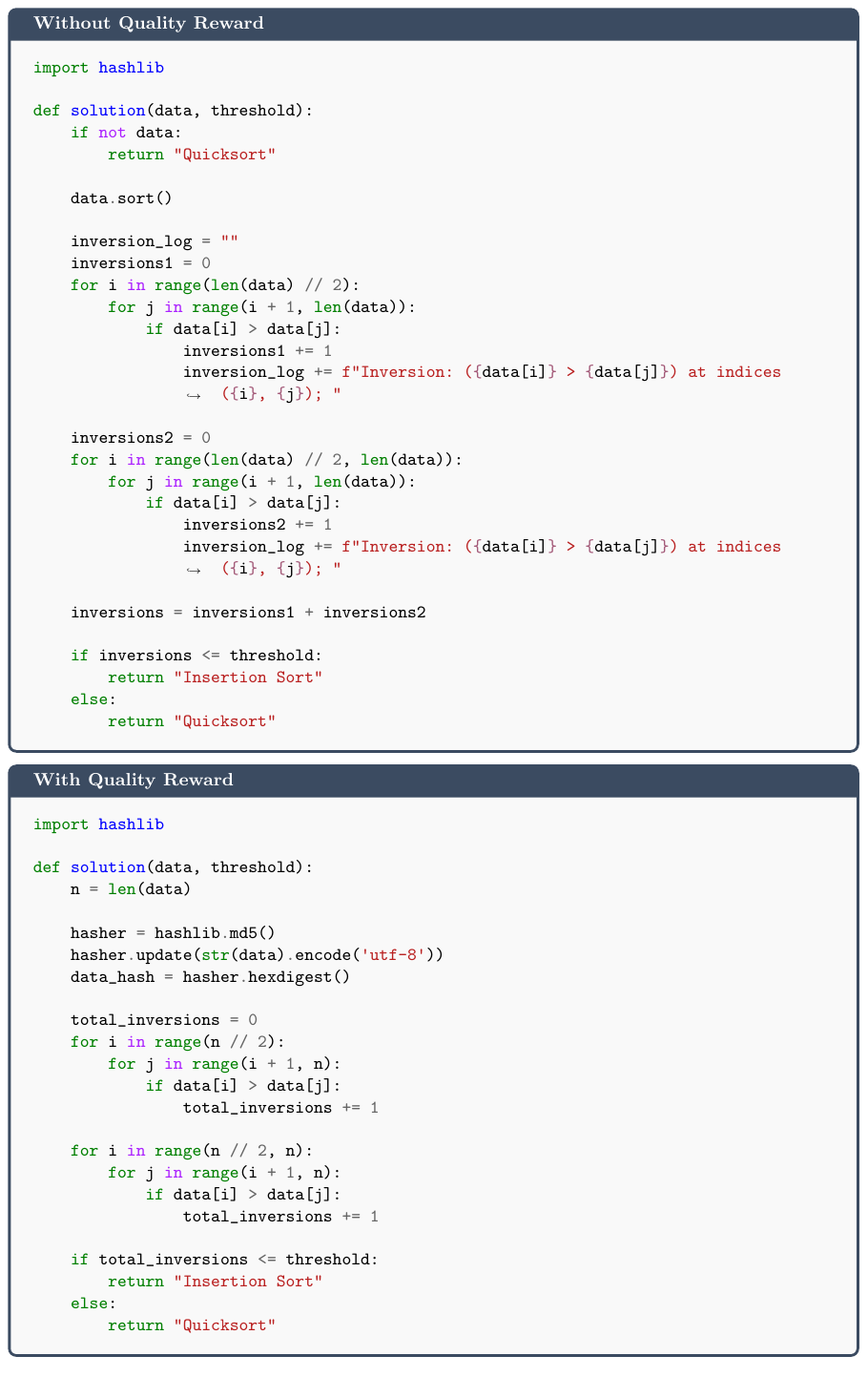}
\end{figure}

\begin{figure}[htbp]
\centering
\includegraphics[width=1.0\linewidth]{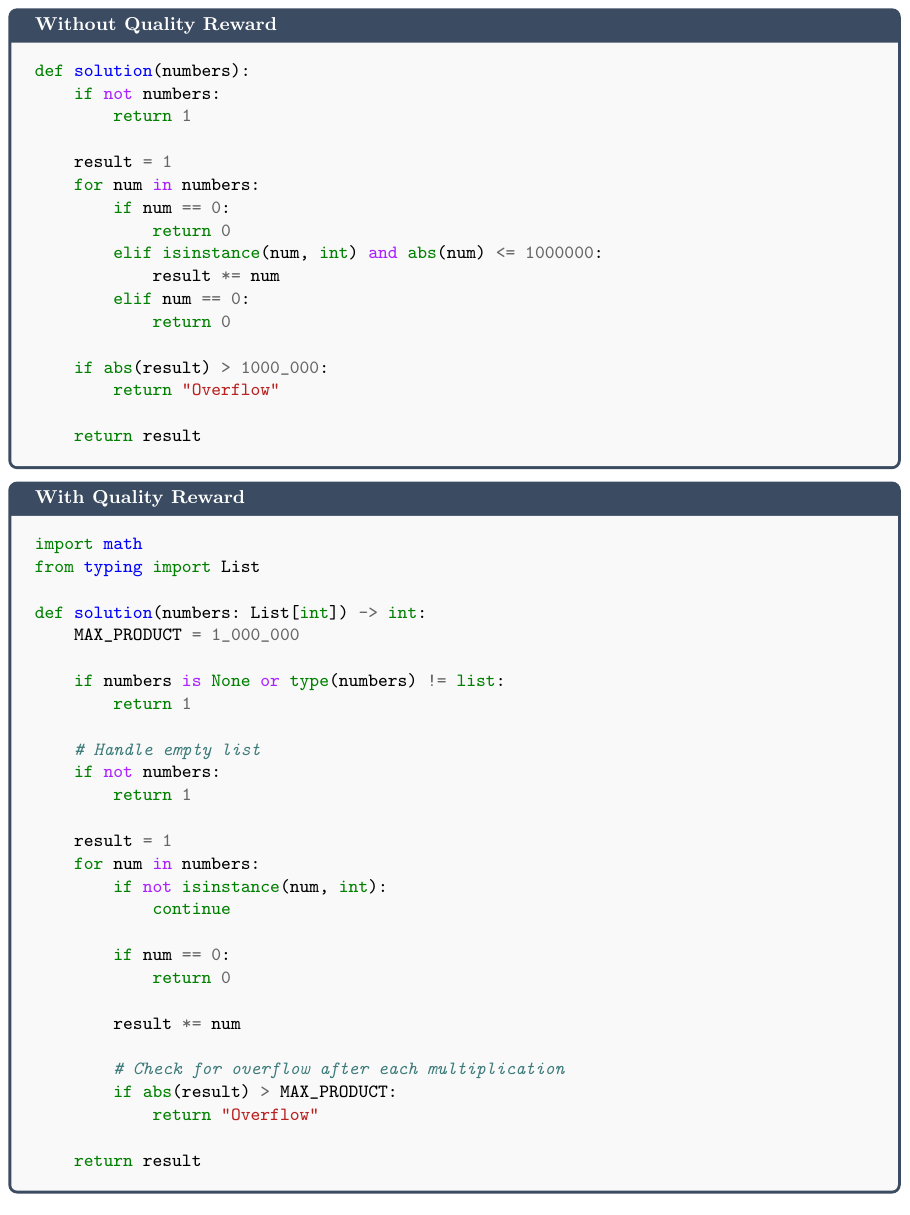}
\end{figure}

%%%%%%%%%%%%%%%%%%%%%%%%%%%%%%%%%%%%%%%%%%%%%%%%%%%%%%%%%%%%

\end{document}